%% file: acl_latex.tex
\pdfoutput=1

\documentclass[11pt]{article}

\usepackage{acl}

\usepackage{times}
\usepackage{latexsym}

\usepackage[T1]{fontenc}

\usepackage[utf8]{inputenc}

\usepackage{microtype}

\usepackage{array}
\usepackage{graphicx}
\usepackage{amsmath}
\usepackage{amsfonts}
\usepackage{bbm}
\usepackage{makecell}
\usepackage{mathtools}

\usepackage{placeins}

\usepackage{tikz}
\newcommand*\circled[1]{\tikz[baseline=(char.base)]{
            \node[shape=circle,draw,inner sep=2pt] (char) {#1};}}

\newcommand{\thickhline}{\noalign{\hrule height 2pt}}

\usepackage{xparse}
\NewDocumentCommand{\heng}
{ mO{} }{\textcolor{red}{\textsuperscript{\textit{Heng}}\textsf{\textbf{\small[#1]}}}}

\NewDocumentCommand{\Carl}
{ mO{} }{\textcolor{blue}{\textsuperscript{\textit{Carl}}\textsf{\textbf{\small[#1]}}}}

\title{Semi-supervised New Event Type Induction and Description via Contrastive Loss-Enforced Batch Attention} 

\author{Carl Edwards \and Heng Ji \\
     University of Illinois Urbana-Champaign \\
     \texttt{\{cne2, hengji\}@illinois.edu}}

\begin{document}
\maketitle

\input{0abstract}

\input{1introduction}

\input{2task}
\input{3method}

\input{4experiment}

\input{5related}

\input{6conclusion}

\nocite{Ando2005,borschinger-johnson-2011-particle,andrew2007scalable,rasooli-tetrault-2015,goodman-etal-2016-noise,harper-2014-learning}

\bibliography{anthology,custom_rebib}
\bibliographystyle{acl_natbib}

\clearpage

\appendix

\input{appendixA}
\input{appendixB}
\input{appendixC}

\input{appendixD}
\input{appendixE}

\input{appendixF}
\input{appendixG}

\input{appendixH}

\input{appendixI}

\end{document}

%% file: 0abstract.tex
\begin{abstract} 
Most event extraction methods have traditionally relied on an annotated set of event types. However, creating event ontologies and annotating supervised training data are expensive and time-consuming. Previous work has proposed semi-supervised approaches which leverage seen (annotated) types to learn how to automatically discover new event types. State-of-the-art methods, both semi-supervised or fully unsupervised, use a form of reconstruction loss on specific tokens in a context. In contrast, we present a novel approach to semi-supervised new event type induction using a masked contrastive loss, which learns similarities between event mentions by enforcing an attention mechanism over the data minibatch. We further disentangle the discovered clusters by approximating the underlying manifolds in the data, which allows us to increase normalized mutual information and Fowlkes-Mallows scores by over 20\% absolute. Building on these clustering results, we extend our approach to two new tasks: predicting the type name of the discovered clusters and linking them to FrameNet frames.\footnote{The programs, data, and resources will be made publicly available for research purposes.} 
\end{abstract}

%% file: 1introduction.tex
\section{Introduction}

Discovering new event types is an important step for adapting information extraction (IE) methods to unseen domains. Existing work \cite{ji2008refining, mcclosky2011event, li2013joint, chen2015event, du2020event, Li2021DocumentLevelEA} traditionally uses a predefined list of event types and their respective annotations to learn an event extraction model. However, these annotations are both expensive and time-consuming to create. This problem is amplified when considering specialization-intensive domains such as scientific literature, which requires years of specialized experience to understand even a specific niche. For example, there are a wide range of otherwise obscure events in biomedical literature \cite{krallinger2017biocreative}, and better IE techniques can empower life-changing breakthroughs in these domains. To adapt IE to these specialized domains, it is critical to discover new event types automatically.

There are two primary approaches in event type induction. The first is completely unsupervised induction. It includes recent neural techniques \cite{huang2016liberal, shen2021corpus}, as well as ad-hoc clustering techniques \cite{sekine2006demand, chambers2011template} and probabilistic generative methods \cite{cheung2013probabilistic, chambers2013event, nguyen2015generative}. The second approach, semi-supervised event type induction, was recently introduced by \citet{huang2020semi}. It proposes leveraging annotations for existing types to learn to discover new types; this enables taking advantage of existing resources. In this work, we pursue the second approach. 



Current state-of-the-art work in event type induction \cite{huang2020semi, shen2021corpus} uses reconstruction-based losses to find clusters of new types. Motivated by recent success in learning representations with contrastive loss \cite{chen2020simple, radford2021learning}, we propose an alternative approach using batch attention and contrastive loss, which achieves state-of-the-art results. Essentially, we consider the attention weight between two event mentions as a learned similarity, and we ensure that the attention mechanism learns to align similar events using a semi-supervised contrastive loss. By doing this, we are able to leverage the large variety of semantic information in pretrained language models for clustering unseen types by using a trained attention head. Unlike \cite{huang2020semi}, we are able to separate clustering from learning, allowing specific task-suited clustering algorithms to be selected.

Batch attention is an attention mechanism taken over a minibatch of samples rather than a sequence. Previous uses of batch attention have been limited. Primarily, it has been used for image classification \cite{cheng2021ba} and satellite imagery \cite{su2019semantic}. In this work, we apply batch attention to natural language instead, which we use for clustering, and we propose the novel idea of enforcing the attention mechanism using contrastive loss. 

To enable our discovered event types to be used in larger IE systems, it is important to extract information regarding the clusters. Previous work has looked to describe clusters---for a given cluster, \citet{huang2016liberal} uses the nearest trigger to the cluster centroid as its name. However, this approach is nebulous and not easily measurable (because the same trigger can correspond to different event types and there is not a quantitative method to determine if the selected trigger defines the cluster well). Instead, we introduce two new information retrieval-styled tasks for type name prediction and FrameNet \cite{baker1998berkeley} frame linking. Type prediction predicts a name for each cluster and is a relatively easy task. FrameNet linking builds on this by linking event types to relevant frames, and is significantly more useful for downstream applications. Our attention-based approach is especially useful here, since it uses the attention mechanism to produce ``clustered'' features which can have auxiliary task-specific losses applied. \\
The major novel contributions of this paper are:

\begin{itemize}
    \item We propose a novel framework for new event type induction which uses contrastive loss to enforce an attention mechanism over the batch. This framework is potentially applicable for semi-supervised clustering and classification problems in other settings where a pretrained model exists (something which is becoming increasingly common).
    \item We show that the base pretrained model selected for event type induction plays a key role in the types which are discovered, since even un-finetuned models rival \citet{huang2020semi}. 
    \item We use the ``clustered'' features produced by our model to extend new event type induction to two novel downstream tasks: type name prediction and FrameNet linking. We show our model with auxiliary losses can improve performance on these tasks.
\end{itemize}


%% file: 2task.tex
\section{Task Descriptions}
\subsection{Semi-supervised Event Type Induction}
We tackle the problem of semi-supervised event type induction, first described by \citet{huang2020semi}. The task is defined as follows: Assume the top 10 most popular event types from the ACE 2005 dataset as defined in \cite{huang2018zero} are known. Given all ACE annotated event mentions, automatically discover the other 23 unseen ACE types. Essentially, this is a semi-supervised clustering task on event mentions. 

\subsection{Downstream Clustering Tasks}
Beyond clustering, we also introduce two new downstream tasks on this problem: type prediction and FrameNet \cite{baker1998berkeley} linking. We structure both of these tasks as information retrieval problems for evaluation. Essentially, given a cluster, one should be able to predict its event type name and to what frame it should be linked. For each cluster, we calculate the most frequent type and consider it to be the ground truth for the cluster. 

\subsubsection{Type Prediction}
For type prediction, the goal is to retrieve the ``name'' of the correct type for a cluster. Thus, we measure Hits@n and mean reciprical rank (MRR), where the corpus consists of the 23 new unseen type names. In practice, we embed the names using our language model and use cosine similarity to the cluster centroid to rank them.

\begin{figure*}[h]
\centering
\includegraphics[width=\textwidth]{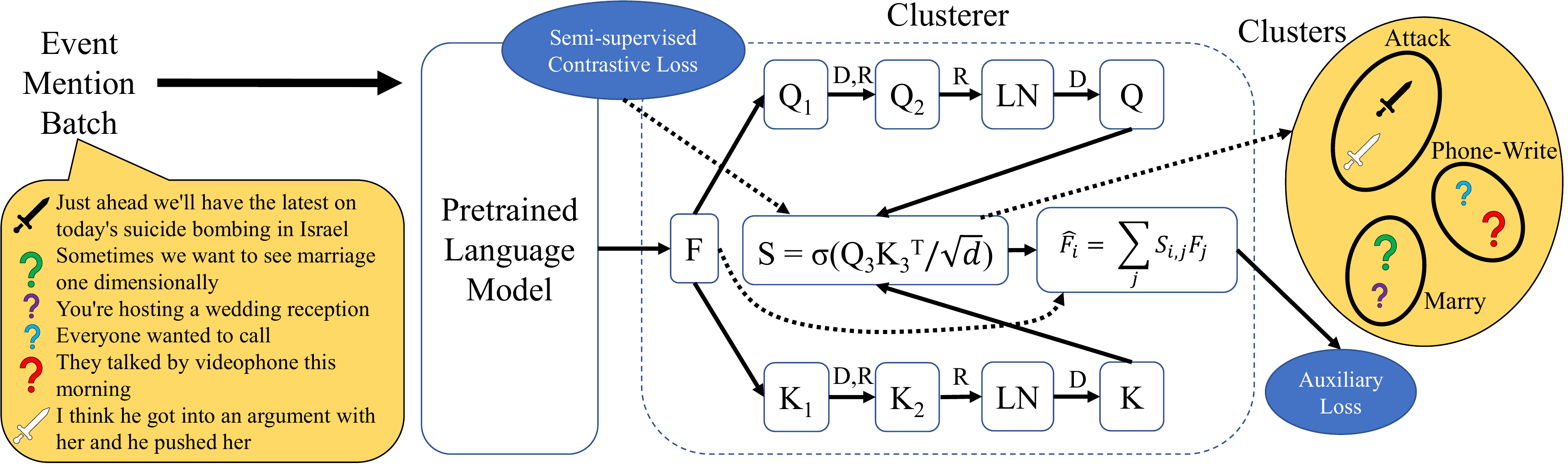}
\caption{Architecture of the proposed approach. Best viewed in color. LN is layer normalization, R is ReLU, and D is dropout. $\sigma$ is softmax for the attention mechanism and sigmoid for the contrastive loss. $\hat{F_i}$ is the clustered features of mention $i$ in the batch. `\textbf{?}' are unseen event types. 
}
\label{fig:architecture}
\end{figure*}

\subsubsection{FrameNet Linking}
FrameNet is the largest event ontology that is publicly available. However, there is not enough annotated training data to train supervised models directly on it. To alleviate this issue, we propose a task linking our newly discovered event types to FrameNet frames. 

For the FrameNet linking, we consider a setup similar to name prediction, where we link clusters to the 1,221 frames in FrameNet 1.7 \cite{ruppenhofer2016framenet}. However, instead of using the type names, we follow \cite{huang2018zero} and manually map the ACE types to one or more frames. The table can be found in Appendix \ref{appendix:linking}. This gives us a mapping into the FrameNet hierarchy. All children of the mapped frames are also considered valid targets. Given an ACE type, we can now link to a set of valid frames. We consider the lowest rank of the valid frames to be the rank of a cluster. In practice, we take the corpus of frame definitions and embed them using our language model.
We then rank them using cosine similarity by comparing to the given cluster centroid. Similarly, this task is measured with Hits@n and MRR. 

%% file: 3method.tex
\section{Methods}

\subsection{Overall Architecture}


Overall, our method, shown in Figure \ref{fig:architecture}, consists of a language model, such as BERT \cite{devlin2018bert}, which produces contextualized representations, followed by a ``clusterer''. Unlike previous work which used specific token embeddings such as triggers \cite{huang2020semi}, we use the sentence where an event occurs as our input. The language model produces an event representation using mean pooling, which is then input into the ``clusterer'' layer. The clusterer layer then produces ``clustered'' features using the attentions (see Section \ref{clusterer}).

\subsection{Back-translation}
Contrastive loss has recently been applied for deep clustering \cite{li2021contrastive, zhong2020deep} and for representation learning \cite{chen2020simple, gao2021simcse,zhang2021pairwise,liu2021simcls}. However, this requires data augmentation to create positive example pairs. For text, some augmentations use back-translation \cite{cao2021cliff, zhang_supporting_2021}. Taking inspiration from these clustering and representation learning techniques, we employ back-translation as data augmentation to create more positive pairs, improving the learning of attention weights between event mentions. 


\subsection{Batch Attention ``Clusterer'' Mechanism} \label{clusterer}

To learn similarities between unseen event mentions, we propose learning an attention mechanism over the stochastic gradient descent minibatch. 
We enforce this attention mechanism using a masked contrastive loss (described in Section \ref{masked_contrastive_loss}). This allows the attention mechanism's behavior to be learned from the seen classes. 

We follow \cite{vaswani2017attention} in implementing a scaled dot product attention, although over the batch instead. Since our ``clusterer'' needs to learn similarities for clustering and then be used for cluster features, we use nonlinear transformations for the query ($Q$) and key ($K$) vectors instead of the linear transformations in \cite{vaswani2017attention}. This nonlinear transformation for $Q$ and $K$ is implemented as a two hidden layer neural network, which is shown in Figure \ref{fig:architecture}. 

Using this attention mechanism, we produce ``clustered features'', which are a convex combination of the different samples from the batch. This allows us to apply an auxiliary loss to the clustered features. We consider this as being analogous to learning on cluster centroids. Specific auxiliary losses can be applied for specific downstream tasks. 

We note that this approach can also be interpreted as a type of feature smoothing, an inner product graph generator, and metric learning.

\subsection{Masked Semi-supervised Contrastive Loss} \label{masked_contrastive_loss}

Recent work, such as CLIP \cite{radford2021learning} and Text2Mol \cite{edwards2021text2mol}, has found great success using contrastive losses between pairs of representations $Q$ and $K$, each $n \times d$ matrices where $n$ is the number of samples of $d$ dimensions.  They obtain the loss $L$ by comparing the product of these matrices ($QK^T$) to a label matrix $Y \in \{0,1\}^{n \times n}$ (which in their case is $Y=I_n$), using cross entropy loss $CE$.

\vspace{-4mm}
$$ L(Q,K) = CE( QK^T, I_n) + CE( KQ^T, I_n)$$

We use a modification of these existing contrastive losses to enforce our batch attention mechanism. We calculate the label matrix as follows. An example is shown in Appendix \ref{appendix:label_example}: Figure \ref{fig:labels_figure}. Given a pair of samples (event mentions) $x_i$ and $x_j$, we consider the pair to be a positive if they are from the same seen event type. We consider the pair to be negative if they are from different seen event types or if one is seen and one unseen. In practice, the labels can be computed using one-hot vectors of the $c$ seen types (the unseen types are zero-hot vectors). These vectors are stacked into a $n \times c$  matrix $O$. The label matrix is computed

\vspace{-2mm}
$$Y = OO^T \lor I_n$$
where $\lor$ is the elementwise logical-or operation.
Following \cite{edwards2021text2mol}, we use binary cross-entropy as the loss between the labels $Y$ and the scaled attention dot products $\frac{QK^T}{\sqrt{d}}$ from \cite{vaswani2017attention}. This gives the following semi-supervised loss:

\vspace{-2mm}
$$ L_{ss}(Q,K) = CE(\frac{QK^T}{\sqrt{d}}, Y)$$
This loss, however, values negative samples much more than positive samples (due to the imbalance). Noticing that once vectors of a negative pair are orthogonal they don't need to be further separated, we introduce a margin $m$. Essentially, we mask out pairs whose dot product is ``too negative'' (in addition to unknown relations between unseen types). This is because the loss would rather optimize the already well-separated negatives instead of the relatively fewer positives. Let $p_{i,j} \in \{0,1\}$ be the label of a pair and $u_{i,j} \in \{0,1\}$ indicate that both $i$ and $j$ are unseen. Our mask, $M$, is calculated

\vspace{-4.2mm}
$$ M_{i,j} = \overline{ \left[ \overline{p_{i,j}}  \land \left( \sigma(\frac{QK^T}{\sqrt{d}})_{i,j} < m \right) \right] \lor u_{i,j} } $$

where $\land$ is elementwise logical-and, $\sigma$ is the sigmoid function, and $\bar{z} $ denotes logical negation of $z$.
This is similarly motivated to the margins used in knowledge graph embedding losses, such as TransE \cite{bordes2013translating}. 
Thus, our loss is: 

\vspace{-1mm}
$$ L_m(Q,K) = M \cdot L_{ss}(Q,K)$$
where in this case we treat $L_{ss}(Q,K)$ as an unreduced loss (so it is a matrix), and $\cdot$ is elementwise multiplication. 

We apply this loss to the query ($Q$) and key ($K$) matrices in the clusterer's batch attention mechanism. We also include the augmented data ($Q'$ and $K'$), giving us a final loss:

\vspace{-5mm}
$$ L_c(Q,K,Q',K') = \sum_{\mathclap{\hat{Q}, \hat{K} \in \{Q, Q'\} \times \{K,K'\}}} L_m(\hat{Q},\hat{K}) $$

\subsection{Auxiliary Loss}
For our downstream tasks, we employ a regression-based auxiliary loss. For each seen instance $x_i$, we maximize the cosine similarity between the clustered features $\hat{F_i}$ and the pretrained language model embedding $B_{t_i}$ of the ground truth type $t_i$ (e.g. the name `attack'). Thus, we get the loss:

\vspace{-4mm}
$$ L_{a}(\hat{F_i}, B_{t_i}, t_i) = 1 - \cos(\hat{F_i}, B_{t_i}) \mathbbm{1}_{\text{seen}}(t_i) $$
where $\mathbbm{1}_{\text{seen}}(t_i)$ indicates whether $t_i$ is a seen type.

\subsection{Stopping Criterion}
\label{stopping_criterion}

For this task, it is not reasonable to use a validation set for stopping. This is because the loss depends only on seen types and their relationships to unseen types. Since the unseen classes are unlabeled and the losses between pairs of unseen are unknown, the model can overfit to the seen data, pushing together clusters of unseen types. We partially address this issue by implementing a margin on negative values, which prevents the model from forcing together unseen clusters as strongly to separate them from seen type events. 
To deal with this issue, we employ unsupervised clustering metrics to decide when to stop training. In particular, we use cosine distance-based silhouette scores to measure the quality of clustering. 
This increases the required compute up to 2x (in practice roughly 1.5x because backpropagation isn't required), but training is already relatively quick, with 10 or less epochs being sufficient. 
We note that this approach can have some variance. To address this issue, we employ a sliding window running average approximation to create a smooth curve of the initial increase and then decrease of the silhouette score. We consider a hybrid approach---we select the window with the highest silhouette score, and then we select the epoch with the highest silhouette score in that window as our stopping point, as shown in Appendix \ref{appendix:early_stopping}.

\subsection{Clustering} 
Any algorithm which can compute clusters from a precomputed distance function can be applied to the learned similarities between event mentions. 
Additionally, we find that the finetuning of the language model by our loss modifies its representations to better form clusters. Thus, this representation can be used in many clustering algorithms as well. 

\subsubsection{Manifold Approximation} \label{manifold_approx}
Inspired by recent work \cite{ros2021team} which uses manifold approximation to interpret large language model-based sentence representations for information retrieval, we incorporate manifold approximation into our clustering approach. To do so, we follow the UMAP \cite{mcinnes2018umap} algorithm to create approximate weights based on estimating neighborhood densities within the data. We calculate these weights using cosine distance as an input, as it has traditionally been effective for language modeling \cite{manning2008, reimers2019sentence}. UMAP attempts to estimate the density by comparing the distance to the k-nearest neighbors. This is used to calculate weights between each pair of data points. Details are given in Appendix \ref{appendix:umap}. Following this, we use agglomerative clustering on the UMAP weights as before.

In our approach, we want to better understand the global clustering landscape, so we use a high value of $k$. In practice, to avoid hyperparameter selection, we set $k$ equal to the size of the data.

%% file: 4experiment.tex
\section{Experimental Results}


Generally, we used default hyperparameters. 
We split the learning rates into BERT and non-BERT parameters following \cite{edwards2021text2mol} with 2e-5 for BERT as in \cite{devlin2018bert} and 1e-4 for other parameters as in \cite{vaswani2017attention}. For the margin parameter, we examined silhouette scores to select 0.5. 

For back-translation, we used four languages, German, French, Spanish, and Chinese, and randomly sampled which language to use for each data point every epoch. We obtained back-translations using the MarianMT translation models \cite{mariannmt}.

For our main experiments, we only use the contrastive loss. We take the average of 5 runs to show that our method consistently outperforms \cite{huang2020semi}. We also calculate clusters using an ensemble of the 5 runs which shows slightly increased performance, which is an expected result in deep neural networks \cite{allen2020towards}. 

\citet{huang2020semi} evaluate these clusters using Geometric NMI, Fowlkes Mallows \cite{fowlkes1983method}, Completeness, Homogeneity, and V-Measure \cite{rosenberg2007v}. We additionally consider adjusted Rand index (ARI) \cite{hubert1985comparing}. In the downstream tasks, given a clustering we also report the average cluster purity and type representation. Given a cluster $i$ of size $n_i$ with most frequent type numbering $n_{f_i}$, purity $p_i = \frac{n_i}{n_{f_i}}$ \cite{manning2008}. Note that this average cluster purity is slightly different than traditional purity; it weights small clusters more which is desirable in our case (like macro vs. micro F1 score). Type representation is the number of unique frequent subtypes, $n_t$, divided by total types, in this case 23. 

\subsection{Language Model}
We select Sentence BERT (SBERT) \cite{reimers2019sentence} as a language model because its pretraining tasks are better suited for clustering than BERT. This is shown in Table \ref{tab:results}, since the clustering from SBERT embeddings can even outperform \cite{huang2020semi} without any semi-supervision. We use a small version of the model\footnote{paraphrase-MiniLM-L12-v2} from HuggingFace \cite{wolf-etal-2020-transformers}, which allows us to use a larger minibatch size of 10. Using larger minibatch sizes is desirable for contrastive loss since the number of negatives scales quadratically with the size. The performance of mini SBERT is notable, as \citet{huang2020semi} used BERT-large, a considerably larger model.

\begin{table*}
\resizebox{\textwidth}{!}{
\centering
\begin{tabular}{ c|c|c|c|c|c|c|c|c }

\multicolumn{2}{c}{\textbf{Method}} & \multicolumn{1}{c}{Clusters} & \multicolumn{1}{c}{Geometric NMI} & \multicolumn{1}{c}{Fowlkes Mallows} & \multicolumn{1}{c}{Completeness} & \multicolumn{1}{c}{Homogeneity} & \multicolumn{1}{c}{V-Measure} & \multicolumn{1}{c}{ARI}  \\
\thickhline
\multicolumn{2}{c|}{One Cluster} & 1 & 0.00 & 25.58 & 100.00 & 0.00 & 0.00 & 0.00 \\
\thickhline
\multicolumn{2}{c|}{SS-VQ-VAE w/o VAE \cite{huang2020semi}} & 500 & 33.45 & 25.54 & 42.76 & 26.17 & 32.47 & - \\
\hline
\multicolumn{2}{c|}{SS-VQ-VAE \cite{huang2020semi}} & 500 & 40.88 & 31.46 & 53.57 & 31.19 & 39.43 & - \\
\thickhline
SBERT & Agglo & 23 & 50.71 & 34.35 & 57.05 & 45.07 & 50.36 & 24.02 \\
\hline
SBERT & Manifold & 23 & 48.75 & 36.02 & 51.32 & 46.30 & 48.68 & 30.21 \\
\thickhline
Attn-Cosine & Agglo & 23 & 46.40 & 34.60 & 49.82 & 43.24 & 46.27 & 26.69 \\
\hline
Attn-DotProduct & Agglo & 23 & 50.17 & 37.48 & 53.50 & 47.06 & 50.06 & 30.13 \\
\hline
Attn & Manifold & 23 & 54.83 & 42.77 & 55.00 & 54.67 & 54.82 & 38.74 \\
\hline
FT-SBERT & Manifold & 23 & 60.28 & 50.63 & 60.19 & 60.37 & 60.28 & 47.24 \\
\thickhline
Attn-DotProduct & Affinity & 49-68 & 56.87 & 35.64 & 49.58 & 65.26 & 56.33 & 30.02 \\
\hline
Attn-Cosine & Affinity & 50-69 & 56.54 & 33.00 & 48.72 & 65.62 & 55.91 & 27.04 \\ 
\thickhline
E-Attn-DotProduct & Agglo & 23 & 56.50 & 43.26 & 59.62 & 53.54 & 56.41 & 37.02 \\
\hline
E-Attn & Agglo & 23 & 59.00 & 46.19 & 58.36 & 59.66 & 59.00 & 42.56 \\
\hline
E-FT-SBERT & Manifold & 23 & \textbf{63.56} & \textbf{52.10} & \textbf{63.11} & 64.01 & \textbf{63.56} & \textbf{48.85} \\
\hline
E-Attn-DotProduct & Affinity & 63 & 60.00 & 38.41 & 51.32 & \textbf{70.15} & 59.28 & 31.78 \\

\end{tabular}
}
\caption{New event type induction results (\%)\footnotemark. The first subcolumn is the input for clustering and the second is the clustering algorithm used. E stands for ensemble and FT for finetuned. SBERT indicates the SBERT representations were used rather than our learned attentions (Attn). Values are the average of 5 runs. Agglo is agglomerative clustering, Affinity is \cite{frey2007clustering}, and Manifold is as described in Section \ref{manifold_approx}. For affinity, each run can produce a slightly different number of clusters.}
\label{tab:results}
\end{table*}

\subsection{Clustering Algorithms}

For clustering, we consider two algorithms which work on precomputed metrics. First, we use agglomerative clustering with average linkage, 
as it tends to be less sensitive to outliers and noisy data \cite{han2011data}. Noise is present in the dataset, often in the form of transcripts (see Section \ref{QCA}). 

We report results following existing clustering literature by using the true number of classes as the cluster number \cite{huang2020deep, li2021contrastive}. In practice it is generally difficult to select the correct number of clusters to use. Due to this, using extra clusters is typically done by previous work \cite{huang2020semi, shen2021corpus}. However, this can inflate the NMI score \cite{nguyen2009information} and benefit qualitative evaluation because of the unbalanced classes in the dataset. As an example, given only 23 clusters (the ground truth), a large class such as `Injure' splits into multiple smaller clusters, which causes rare event types to be merged. Results show that 19 / 23 types are represented by a cluster in the 50 cluster case versus only 16 / 23 in the 23 cluster case. This makes results appear better for more clusters. Silhouette scores are higher for 23 clusters, however. 

Unlike existing work \cite{huang2020semi}, the number of clusters is unimportant for our learning process and can be selected afterwords, such as by selecting a high number as in \cite{huang2020semi, shen2021corpus} or automatically with affinity propagation \cite{frey2007clustering}. Affinity propagation selects exemplars to automatically determine the number of clusters. Our approach is especially useful here, since affinity propagation does not complete when applied to default SBERT representations but does when using our contrastive loss-enforced attentions.




\subsection{Results}

We compare our results with \citet{huang2020semi}, who first introduced this task, in Table \ref{tab:results}. We find that just our choice of language model outperforms the baseline. 
Also, using dot products is more effective for our learned attention metric than cosine distance, since dot product without normalization, as in our attention mechanism, indicates confidence of clustering a pair of samples together (This is because the contrastive loss uses sigmoid). 

\footnotetext{\cite{huang2020semi} appears to have used the former scikit-learn default of geometric NMI, which is why their v-score doesn't equal arithmetic NMI.}

\subsubsection{Manifold Approximation}
We find manifold approximation to be very effective in our experiments. Intuitively, we understand this manifold approximation as untangling the cluster manifolds from each other in the high-dimensional representation space. 
Interestingly, the results using the finetuned SBERT representations perform better than the results on the learned similarities. We find this to be quite interesting, especially because the representations change an average of 0.6 cosine distance from their starting points, as shown in Appendix \ref{appendix:SBERT_change}. Our method causes SBERT to inherently learn representations more amenable for clustering.

While manifold approximation works well for clustering here, we note that using UMAP for clustering is considered controversial.\footnotemark \text{ }While it works well in many cases, there are potential issues with artifacts or false tearing of clusters. We leave analysis of the interaction between high-dimensional semantic spaces obtained from language models and manifold approximation to future work.

\footnotetext{\href{https://umap-learn.readthedocs.io/en/latest/clustering.html}{https://umap-learn.readthedocs.io/en/latest/clustering.html}}

\subsection{Qualitative Cluster Analysis} \label{QCA}

\begin{table}
\resizebox{\columnwidth}{!}{
\centering
\begin{tabular}{ c|c }
 Cluster Strength &  Clusters \\
\hline
 \thead{Very Strong \\ (> 80\% Purity)} & \thead{Injure, Sue, Phone-Write,\\ Declare-Bankruptcy, Demonstrate, Trial-Hearing} \\
 \hline
 \thead{Strong \\ (60-80\% Purity)} & \thead{Be-Born, Start-Position, \\ Charge-Indict, Marry} \\
 \hline
 \thead{Ok \\ (40-60\% Purity)} & \thead{Release-Parole, Appeal, Injure} \\
 \hline
 \thead{Mixed \\ (20-40\% Purity)} & \thead{Convict, Fine, Trial-Hearing, \\ Start-Org, Start-Position, Charge-Indict} \\
 \thickhline
 \thead{Small Clusters \\ (< 2 samples)} & \thead{Trial-Hearing, Nominate, \\ Start-Position, Phone-Write} \\
\end{tabular}
}
\caption{Clusters sorted into purity classes.}
\label{tab:cluster_class}
\end{table}

\begin{figure*}[h]
\centering
\includegraphics[width=\textwidth]{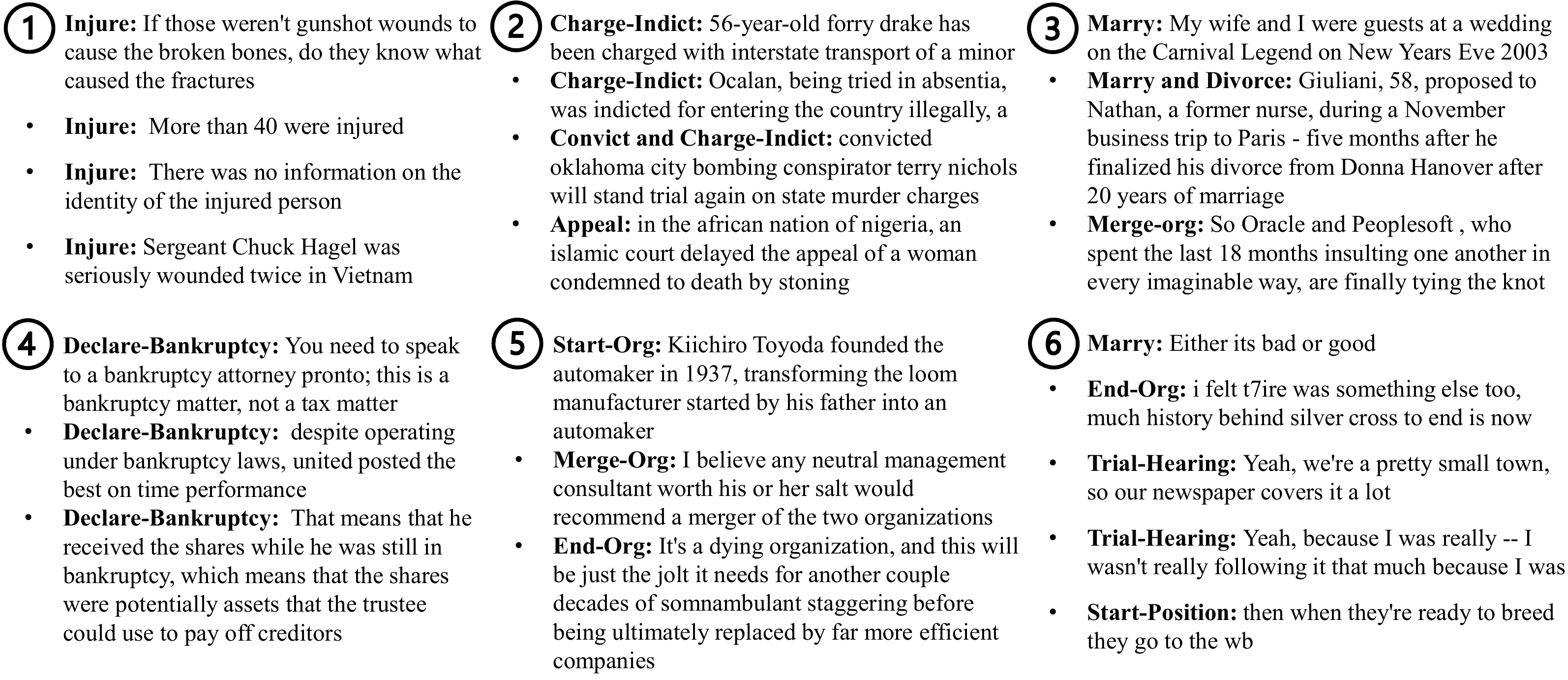}
\caption{Cluster Examples: Injure, Charge-Indict, Marry, Bankruptcy, Start-Org, and Bad Data, respectively. 
}
\label{fig:examples}
\end{figure*}

We analyze the clusters produced by our best result, the ensemble. We classify the clusters according to purity in Table \ref{tab:cluster_class}. We show examples from numbered clusters in Figure \ref{fig:examples}. Certain types of clusters, such as Injure \circled{1} and Demonstrate \circled{4}, form very strong clusters. We believe this is likely related to their size and lack of overlap with other types. There are two common sources of error: the first is semantic overlap. Start-org, merge-org, and end-org tend to overlap \circled{5}. Marry and divorce also slightly overlap \circled{3}---in the 23 cluster case they merge into one cluster, but in the 50 cluster case they are separate. Most types of courtroom related events---Charge-Indict, Trial-Hearing, Convict, Release-Parole, Appeal, Execute, Acquit, Extradite---have some degree of overlap \circled{2}. Second, the other main source of errors is ``duplicates''. This occurs in our method because two or more events can occur in the same event mention \circled{2}, \circled{3}. Since our method does not account for triggers, it cannot distinguish between duplicate mentions with multiple triggers. Future work can address this issue by combining our method with an existing trigger-based method such as \cite{huang2020semi}.
We also find that our method clusters ``junk'' data together \circled{6}, which are usually from transcripts. Errors occasionally occur from metaphorical language, such as when companies are ``married'' \circled{3}. We show more detailed examples of these observations in Appendix \ref{appendix:cluster_example}.

\subsection{Downstream Tasks}

\begin{table*}
\resizebox{\textwidth}{!}{
\centering
\begin{tabular}{ c|c|c|c|c|c|c|c|c|c }
 \multicolumn{1}{c}{Method} & \multicolumn{1}{c}{Mean Rank} & \multicolumn{1}{c}{Hits@1} & \multicolumn{1}{c}{Hits@3} & \multicolumn{1}{c}{Hits@5} & \multicolumn{1}{c}{Hits@10} & \multicolumn{1}{c}{Hits@15} & \multicolumn{1}{c}{MRR} & \multicolumn{1}{c}{Average Purity} & \multicolumn{1}{c}{Type Representation} \\
\hline
 Default-23 & 5.17 & 34.8\% & 47.8\% & 60.9\% & 82.6\% & 100\% & 0.477 & 25\% & 47.8\% \\
\hline
 FT-23 & 4.43 & 56.5\% & 65.2\% & 78.2\% & 82.6\% & 91.3\% & 0.660 & 58.9\% & 65.2\% \\
\thickhline
 E-Default-23 & 3.65 & 60.9\% & 69.6\% & 69.6\% & 95.7\% & 100\% & 0.679 & 68.6\% & 69.6\% \\
\hline
 E-FT-23 & 5.13 & 56.5\% & 65.2\% & 69.6\% & 87.0\% & 87.0\% & 0.650 & 68.6\% & 69.6\% \\
\hline
 E-Default-50 & 4.40 & 56.0\% & 60.0\% & 68.0\% & 90.0\% & 96.0\% & 0.630 & 69.3\% & 82.6\% \\
\thickhline
 Perfect-Default-23 & 2.30 & 69.6\% & 73.9\% & 82.6\% & 95.7\% & 100\% & 0.758 & 100\% & 100\% \\
 \hline
 Perfect-FT-23 & 2.83 & 73.9\% & 82.6\% & 91.3\% & 91.3\% & 95.7\% & 0.800 & 100\% & 100\% \\
\end{tabular}
}

\caption{Results for cluster to name prediction task. Default indicates SBERT representations are used to compute cluster centroids. FT indicates finetuned SBERT using our contrastive auxiliary loss instead. x is the number of clusters in the clustering. E indicates that the ensemble clustering is used instead. Perfect indicates the ground truth clustering. Type representation shows the percent of unseen types representing the majority of a cluster.}
\label{tab:type_pred}
\end{table*}

\begin{table*}
\resizebox{\textwidth}{!}{
\centering
\begin{tabular}{ c|c|c|c|c|c|c|c|c|c }
 \multicolumn{1}{c}{Method} & \multicolumn{1}{c}{Mean Rank} & \multicolumn{1}{c}{Hits@1} & \multicolumn{1}{c}{Hits@5} & \multicolumn{1}{c}{Hits@10} & \multicolumn{1}{c}{Hits@50} & \multicolumn{1}{c}{Hits@100} & \multicolumn{1}{c}{MRR} & \multicolumn{1}{c}{Average Purity} & \multicolumn{1}{c}{Type Representation} \\
\hline
 Default-23 & 95.9 & 4.3\% & 21.7\% & 26.1\% & 30.4\% & 34.8\% & 0.128 & 25\% & 47.8\% \\
\hline
 FT-23 & 156.9 & 30.4\% & 30.4\% & 34.8\% & 43.5\% & 47.8\% & 0.336 & 57.4\% & 65.2\% \\
\thickhline
 E-Default-23 & 72.7 & 17.4\% & 30.4\% & 39.1\% & 47.8\% & 65.2\% & 0.264 & 68.6\% & 69.6\% \\
\hline
 E-FT-23 & 115.7 & 21.7\% & 34.8\% & 34.8\% & 43.5\% & 52.2\% & 0.308 & 68.6\% & 69.6\% \\
\thickhline
 Perfect-Default-23 & 15.9 & 26.1\% & 39.1\% & 52.2\% & 65.2\% & 73.9\% & 0.374 & 100\% & 100\% \\
 Perfect-FT-23 & 42.7 & 47.8\% & 56.5\% & 60.9\% & 69.6\% & 69.6\% & 0.539 & 100\% & 100\% \\
\end{tabular}
}
\caption{Results for cluster to frame linking task. See Table \ref{tab:type_pred} for notation.}
\label{tab:framenet}
\end{table*}

For the downstream tasks, we use different clusterings and try to discover information about the clusters. As a baseline, we compare against default (not finetuned) SBERT clustering and ground truth (perfect) clusters. We compare these to our ensemble clustering. For type prediction, we use default SBERT embeddings to compute cluster centroids and then compare to the SBERT representation of the type name (e.g. `injure'). For FrameNet linking, we use the frame definition instead of the name (e.g. ``The words [...] describe situations in which an Agent or a Cause injures a Victim [...]''). We also use an auxiliary loss, $L_a$, which we apply to a 1-layer neural network on the clustered features $\hat{F}$. This extra layer is employed to allow multiple auxilliary losses: we leave those experiments for future work. We compare using these finetuned representations in addition to default SBERT. Results are shown in Tables \ref{tab:type_pred} and \ref{tab:framenet}.

We find that our ensemble clustering outperforms the default SBERT clustering, and that we are able to recover the event type $60\%$ of the time. For the ground truth clusters, our finetuning with an auxiliary loss improves MRR and Hits@1 over the default SBERT representations. Frame linking is much more difficult, since there are 1,221 frames, but we are able to recover the correct frame for $30\%$ of clusters, while default SBERT only achieves $4\%$. Notably, the auxiliary loss clustering (FT-23) even outperforms our ensemble clustering, demonstrating the flexibility of our model architecture. Using perfect clustering, our finetuned model achieves nearly 50\% Hits@1, doubling the performance of the default SBERT model. 
The finetuning loss allows the model to train on the combined cluster features, which approximates the centroid of a cluster. This promotes the potential cluster to be shaped so that its centroid is better suited for the downstream tasks--being most similar to the correct representation of the task name and description.

%% file: 5related.tex
\section{Related Work}

Although event extraction has long been studied \cite{grishman1997information, ji2008refining, mcclosky2011event, li2013joint, chen2015event, du2020event, Li2021DocumentLevelEA}, recent focus has turned towards discovering events without annotations. It includes recent neural techniques \cite{huang2016liberal, liu2019open, shen2021corpus}, as well as ad-hoc clustering techniques \cite{sekine2006demand, chambers2011template, yuan2018open} and probabalistic generative methods \cite{cheung2013probabilistic, chambers2013event, nguyen2015generative}. Semi-supervised event type induction was recently introduced by \citet{huang2020semi}. Zero-shot event extraction frameworks, such as \cite{huang2018zero}, can be used to perform event extraction on the newly discovered types. 

Several new unsupervised deep clustering approaches use contrastive loss for clustering images \cite{li2021contrastive, zhong2020deep} and text \cite{zhang_supporting_2021}. These methods require data augmentation to create positive example pairs. Contrastive loss has also been applied to learn representations. SimCLR \cite{chen2020simple, chen2020improved} uses image augmentations for unsupervised representation learning. Follow-up work has applied this loss to natural language \cite{gao2021simcse,zhang2021pairwise,liu2021simcls}, with some augmentations being back-translated text \cite{cao2021cliff}.
\citet{gunel2020supervised} use fully supervised contrastive loss to finetune language models. 

Batch attention has been investigated a little in the literature, such as for satellite imagery prediction \cite{su2019semantic} or image classification \cite{cheng2021ba}; however, it has not been used to learn clustered features. \citet{seidenschwarz2021learning} recently proposed a related idea for a message-passing network weighted by attention for clustering images, which is probably the most related idea to ours. We instead directly consider (contrastive loss-enforced) attention weights for clustering.

Semi-supervised clustering is a relatively understudied problem compared to semi-supervised classification \cite{van2020survey}. \citet{bair2013semi} summarizes several methods, most of which are based on k-means.

%% file: 6conclusion.tex
\section{Conclusion and Future Work}

In this work, we present an exciting new approach for event type induction, where we use contrastive loss to control the learning of a batch attention mechanism for both finding and learning about new cluster types. We also consider manifold approximation for clustering, and we introduce two new downstream tasks: name prediction and FrameNet linking. This new approach opens several interesting problems for future work. First, this method can potentially be incorporated with reconstruction loss-based approaches, which might improve results or obviate the early stopping criterion. Alternatively, the stopping criterion can be integrated into a loss function for better stopping control. It is notable that this would enable a two-step process of learning clusters and then performing knowledge distillation using those clusters (or an ensemble) while also learning other desired losses. Future work can investigate the interaction of manifold approximation with large language models and integrate it directly into the clusterer subnetwork. Finally, the FrameNet linking task can be extended to Wikidata Q-Node linking, which contains millions of nodes. 
Our approach may also be applicable in other modalities with strong pretrained models, such as for semi-supervised image clustering. 

%% file: appendixA.tex
\FloatBarrier
\section{How much do SBERT representations change?}
\label{appendix:SBERT_change}

\begin{figure}[h]
\centering
\includegraphics[width=\linewidth]{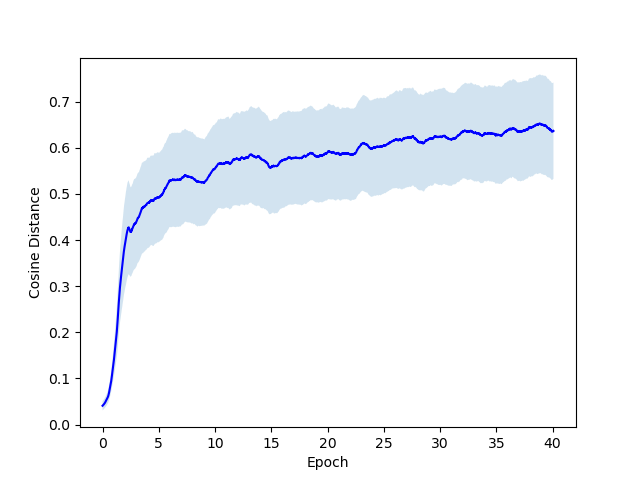}
\caption{Change in SBERT representations from original representation of inputs. This shows that the representations change significantly from their starting point during finetuning. Shaded area is one standard deviation. 
}
\label{fig:rep_change}
\end{figure}

%% file: appendixB.tex
\FloatBarrier
\section{Label Matrix Example}
\label{appendix:label_example}

\begin{figure}[h!]
\centering
\includegraphics[width=\linewidth]{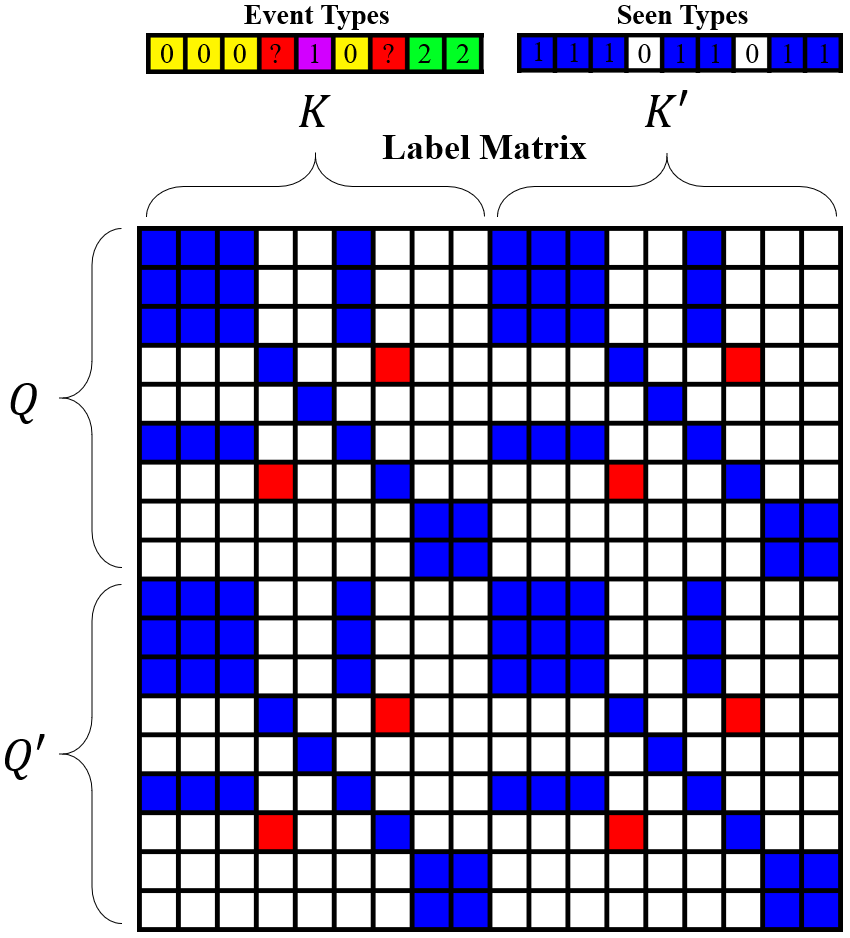}
\caption{Best viewed in color. Visualization of the label matrix $Y$ used in the loss. Blue is positive, white is negative, and red is masked. Note that the mask for the negatives less than the margin is not shown. 
The event types and corresponding ``seen'' boolean vector are also shown, and are used to construct the label matrix. $Q$ and $K$ are corresponding queries and keys to the labels, while $Q'$ and $K'$ are augmented data.
}
\label{fig:labels_figure}
\end{figure}

%% file: appendixC.tex
\FloatBarrier

\section{Manual ACE05 to FrameNet Linking}
\label{appendix:linking}

\begin{table}[!htb]
\resizebox{\columnwidth}{!}{
\centering
\begin{tabular}{ c|c }
 ACE Type &  Frame \\
\thickhline
Appeal & Appeal \\
Be-Born & Birth\_scenario \\
Charge-Indict & Notification\_of\_charges \\
Convict & Verdict \\
Declare-Bankruptcy & Wealthiness \\
Demonstrate & Protest \\
Divorce & Personal\_relationship \\
End-Org & Organization  |  Process\_end \\
Extradite & Extradition \\
Fine & Fining \\
Injure & Cause\_harm | Experience\_bodily\_harm \\
Marry & Forming\_relationships \\
Nominate & Appointing \\
Phone-Write & Contacting \\
Release-Parole & Releasing\_from\_custody \\
Start-Org & Organization | Process\_start \\
Start-Position & Being\_employed | Process\_start \\
Sue & Judgment\_communication \\
Trial-Hearing & Trial \\
Pardon & Pardon \\
Merge-Org & Organization | Amalgamation \\
Acquit & Verdict \\
Execute & Execution \\
Attack & Attack \\
Transport & Transportation\_status \\
Die & Death \\
Meet & Make\_acquaintance | Meet\_with | Come\_together \\
Arrest-Jail & Arrest | Prison | Imprisonment | Being\_incarcerated \\
Sentence & Sentencing \\
Transfer-Money & Commerce\_money-transfer \\
Elect & Change\_of\_leadership | Choosing \\
Transfer-Ownership & Commerce\_goods-transfer \\
End-Position & Being\_employed | Process\_end \\
\end{tabular}
}
\caption{Mapping from ACE types to FrameNet frames. Some ACE types required multiple frames to be correctly mapped, which is indicated by `` | ''.}
\label{tab:ACE_framenet_mapping}
\end{table}

%% file: appendixD.tex
\FloatBarrier
\section{Visualization}
\label{appendix:visualization}

\begin{figure*}
\centering
\includegraphics[width=\textwidth]{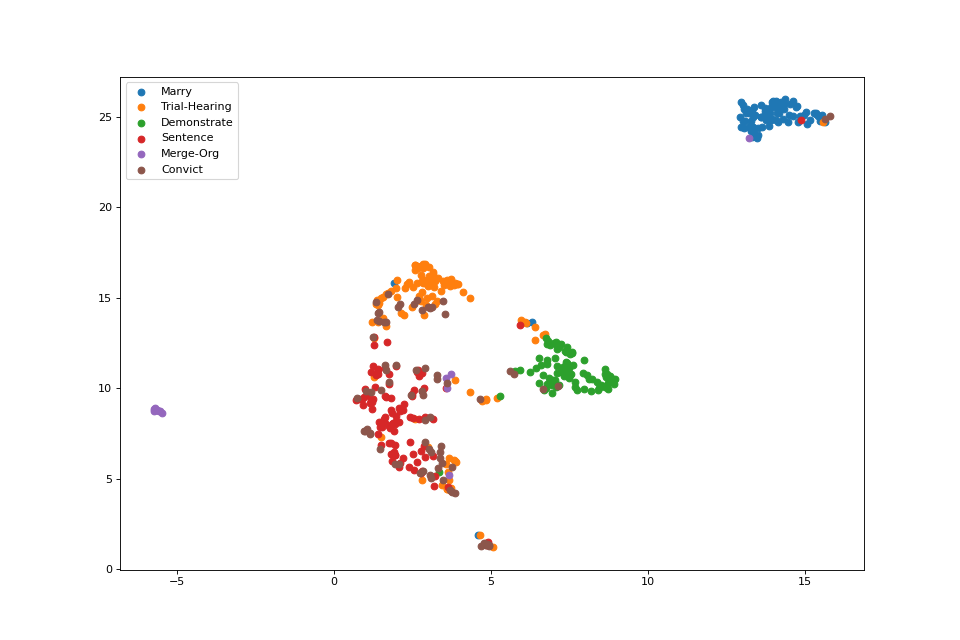}
\caption{Visualization following \cite{huang2020semi} for one of the runs.\footnotemark Note that our clusters have much less errors. 
}
\label{fig:lifu_vis}
\end{figure*}

We visualize unseen event mentions using UMAP \cite{mcinnes2018umap} given a precomputed distance matrix of the cosine distance between $Q$ and $K$. Following \cite{huang2020semi}, we show the results on six unseen types in Figure \ref{fig:lifu_vis}. Sentence and convict overlap significantly, which makes intuitive sense as they are semantically very similar. Unlike \cite{huang2020semi}, trial-hearing forms its own cluster. 

\begin{figure*}
\centering
\includegraphics[width=\textwidth]{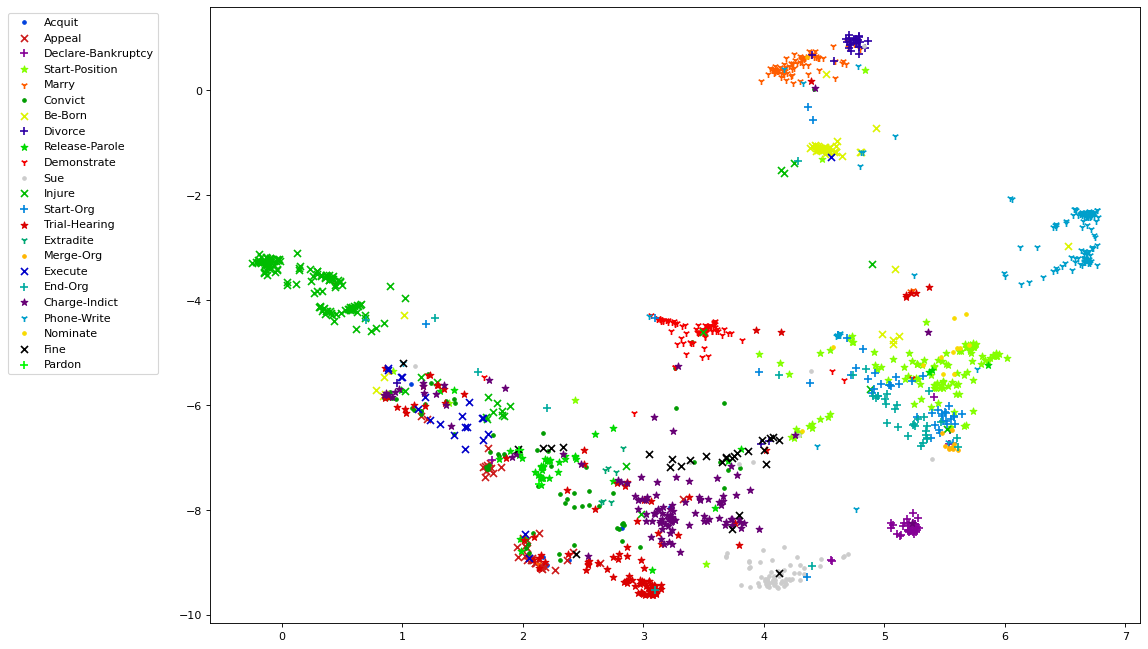}
\caption{Visualization of all unseen types as seen by manifold approximation. Note that dimensionality reduction to 2D renders it difficult to understand with this high number of clusters, but the overall semantics of the space are interesting. 
}
\label{fig:allunseen}
\end{figure*}

\footnotetext{Note that there is a mistake in \cite{huang2020semi}, since ``sentence'' is a seen type in \cite{huang2018zero}}

%% file: appendixE.tex
\FloatBarrier
\section{Cluster Examples}
\label{appendix:cluster_example}

We show extensive examples of our noted observations in Tables \ref{tab:cluster_example1} and \ref{tab:cluster_example2}. Namely, start-org, merge-org, and end-org tend to overlap. Marry and divorce slightly overlap in the 23 cluster case. Most types of courtroom related events---Charge-Indict, Trial-Hearing, Convict, Release-Parole, Appeal, Execute, Acquit, Extradite---have some degree of overlap. There are ``duplicates'' when two or more events can occur in the same event mention. 
We also note the cluster of ``junk'' data, where the label isn't obvious from the event mention.

\begin{table*}
\resizebox{\textwidth}{!}{
\centering
\begin{tabular}{ c|c|m{12cm} }

\multicolumn{1}{c}{Cluster Type} & \multicolumn{1}{c}{Purity} & \multicolumn{1}{c}{Cluster Member Example Types and Inputs} \\
\thickhline\
Injure & 98.3\% & 
\begin{itemize} 
    \item \textbf{Injure:} According to other reports reaching here, five Syrian bus passengers were killed and 10 others were injured on Sunday morning when a US missile hit the bus they were traveling in near the Iraqi border 
    \item \textbf{Injure:}  More than 40 were injured 
    \item \textbf{Injure:}  There was no information on the identity of the injured person
\end{itemize} \\
\hline
Declare-Bankruptcy & 95.0\% & 
\begin{itemize} 
    \item \textbf{Declare-Bankruptcy:} You need to speak to a bankruptcy attorney pronto; this is a bankruptcy matter, not a tax matter 
    \item \textbf{Declare-Bankruptcy:}  despite operating under bankruptcy laws, united posted the best on time performance 
    \item \textbf{Declare-Bankruptcy:}  That means that he received the shares while he was still in bankruptcy, which means that the shares were potentially assets that the trustee could use to pay off creditors
\end{itemize} \\
\hline
Demonstrate & 95.0\% & 
\begin{itemize} 
    \item \textbf{Demonstrate:} The protest follows a string of others involving tens of thousands of peace activists across Japan since January
    \item \textbf{Demonstrate:}  No, I don't demonstrate against anybody during a war 
    \item \textbf{Demonstrate:}  Several thousand demonstrators also gathered outside the White House in Washington, accompanied by a major security presence
\end{itemize} \\
\hline
Charge-Indict & 64.4\% & 
\begin{itemize} 
    \item \textbf{Charge-Indict:} 56-year-old forry drake has been charged with interstate transport of a minor 
    \item \textbf{Charge-Indict:}  Ocalan, being tried in absentia, was indicted for entering the country illegally, a misdemeanor 
    \item \textbf{Convict and Charge-Indict:}  convicted oklahoma city bombing conspirator terry nichols will stand trial again on state murder charges
    \item \textbf{Appeal:} in the african nation of nigeria, an islamic court delayed the appeal of a woman condemned to death by stoning
\end{itemize} \\
\hline
Start-Position & 64.4\% & 
\begin{itemize} 
    \item \textbf{Start-Position:} Many Iraqis boycotted the meeting in opposition to U.S. plans to install Garner atop an interim administration
    \item \textbf{Start-Position:}  The meeting was Shalom's first encounter with an Arab counterpart since he took office as Israel's foreign minister on February 27
    \item \textbf{Start-Org:}  Meeting in the biblical birthplace of the prophet Abraham, delegates from Iraq's many factions discussed the role of religion in the future government and ways to rebuild the country
\end{itemize} \\
\end{tabular}
}
\caption{Examples of discovered clusters. Charge-Indict shows an example of a duplicate---an input with multiple event types. It also shows how courtroom related events can overlap. For Start-Position, there are some errors related to the Middle East, which occurs frequently in the Start-Position mentions. }
\label{tab:cluster_example1}
\end{table*}

\begin{table*}
\resizebox{\textwidth}{!}{
\centering
\begin{tabular}{ c|c|m{12cm} }
\multicolumn{1}{c}{Cluster Type} & \multicolumn{1}{c}{Purity} & \multicolumn{1}{c}{Cluster Member Example Types and Inputs} \\
\thickhline\
Marry & 70.2\% & 
\begin{itemize} 
    \item \textbf{Marry:} My wife and I were guests at a wedding on the Carnival Legend on New Years Eve 2003
    \item \textbf{Marry and Divorce:}  Giuliani, 58, proposed to Nathan, a former nurse, during a November business trip to Paris - five months after he finalized his divorce from Donna Hanover after 20 years of marriage 
    \item \textbf{Phone-Write:}  All the guests were folks who had met the bride and groom (an attractive young couple who were sailing alone) virtually on cruisecritic
\end{itemize} \\
\hline
Start-Org & 34.7\% & 
\begin{itemize} 
    \item \textbf{Start-Org:} Kiichiro Toyoda founded the automaker in 1937, transforming the loom manufacturer started by his father into an automaker
    \item \textbf{Merge-Org:}  I believe any neutral management consultant worth his or her salt would recommend a merger of the two organizations 
    \item \textbf{End-Org:}  It's a dying organization, and this will be just the jolt it needs for another couple decades of somnambulant staggering before being ultimately replaced by far more efficient companies
\end{itemize} \\
\hline
Bad Data & - & 
\begin{itemize} 
    \item \textbf{Marry:} Either its bad or good
    \item \textbf{End-Org:}  i felt t7ire was something else too, much history behind silver cross to end is now 
    \item \textbf{Trial-Hearing:}  Yeah, we're a pretty small town, so our newspaper covers it a lot
\end{itemize} \\
\hline
Phone-Write & 86.2\% & 
\begin{itemize} 
    \item \textbf{Phone-Write:} Let's see, my first call I got was from Russia
    \item \textbf{Phone-Write:}  I'm chewing gum and talking on the phone while writing this note
    \item \textbf{Phone-Write:}  He wants to call his mom in Houston
\end{itemize} \\
\hline
Sue & 92.5\% & 
\begin{itemize} 
    \item \textbf{Sue:} Buyers and sellers also would have to agree not to pursue further cases in foreign courts
    \item \textbf{Sue:}  The cost of class actions is factored into the cost of everything you buy
    \item \textbf{Sue:}  The average number of suits against a neurosurgeon is five in South Florida
\end{itemize} \\
\end{tabular}
}
\caption{More examples of discovered clusters. Start-Org shows the semantic overlap between the organization-related clusters. Bad Data shows a cluster which mostly contains unclear input. }
\label{tab:cluster_example2}
\end{table*}

%% file: appendixF.tex
\FloatBarrier
\section{Early Stopping Example}
\label{appendix:early_stopping}

\begin{figure}[h]
\centering
\includegraphics[width=\linewidth]{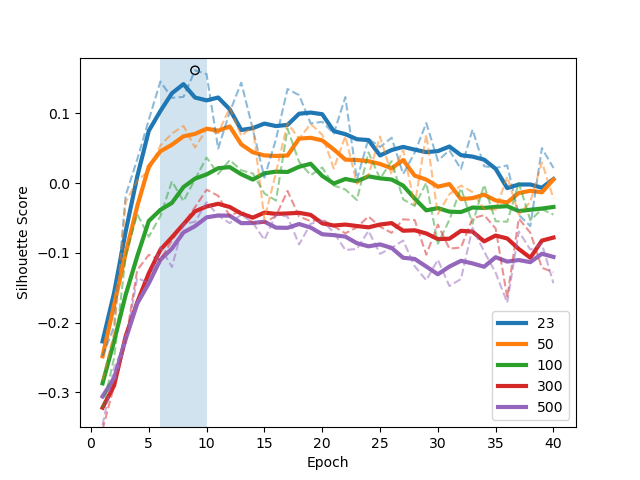}
\caption{Bold lines are sliding window averages of size 5 over silhouette scores. Dotted lines are unsmoothed scores. Legend shows number of clusters. Note that the silhouette scores initially increase and then decay as overfitting occurs, resulting in the need for early stopping. Here, for 23 clusters, epoch 8 has the highest average score. The blue region shows the window around it, and epoch 9 (the black dot) is selected for stopping. 
}
\label{fig:early_stopping}
\end{figure}

%% file: appendixG.tex
\section{Evaluation Metrics}

For the information retrieval metrics, given a list of rankings $R$,

$$ Mean Rank = \frac{1}{n}\sum_{i=1}^n R_i $$
$$ MRR = \frac{1}{n}\sum_{i=1}^n \frac{1}{R_i} $$
$$ Hits@m = \frac{1}{n}\sum_{i=1}^n \mathbbm{1}_{R_i \leq m} $$

\subsection{Clustering Evaluation Metrics}
\label{appendix:clustering_metrics}
Assume there are two clusterings: a set of (ground truth) classes $C$ and a set of (predicted) clusters $K$. Each have $N$ samples. Denote $TP$ as true positives, the number of data point pairs that are in the same cluster in $C$ and $K$. $FP$ is the false positives, the number of data point pairs that belong in the same cluster in $C$ but are not in $K$. $FN$ is false negatives, the number of data point pairs that are in the same cluster $K$ but not in the same ground truth cluster in $C$. $TN$ is the number of data point pairs that are in different clusters in both $C$ and $K$. 

\begin{itemize}
    \item \textbf{Geometric NMI} is the normalized mutual information between two cluster assignments. It is defined:
    
    $$ NMI = \frac{I(C, K)}{mean (H(C), H(K))} $$
    where I is the mutual information and H is entropy. In this case, $mean$ is the geometric mean.
    
    $$ mean(x_1, ..., x_n) = \left( \prod_{i=1}^n x_n \right)^\frac{1}{n} $$
    We note that arithmetic NMI using the arithmetic mean is often reported, but that it is equivalent to V-Measure. 
    \item \textbf{Fowlkes Mallows} \cite{fowlkes1983method} is used to evaluate the similarity between a clustering and the ground truth. It is the geometric mean of pairwise precision and recall. 
    
    $$ FM = \frac{TP}{\sqrt{(TP+FP)(TP+FN)}} $$

    \item \textbf{Completeness} \cite{rosenberg2007v}
    Completeness measures whether all of the data points assigned to a single class are assigned to a single cluster. It is defined:
    \[
      c =
      \begin{cases}
                                       1 & \text{if $H(K,C) = 0$} \\
                                       1 - \frac{H(K|C)}{H(K)} & \text{else} 
      \end{cases}
    \]
    \item \textbf{Homogeneity} \cite{rosenberg2007v} measures whether data points in a cluster are all assigned the the same class. It is symmetric to completeness:

    \[
      h =
      \begin{cases}
                                       1 & \text{if $H(C,K) = 0$} \\
                                       1 - \frac{H(C|K)}{H(C)} & \text{else} 
      \end{cases}
    \]
    \item \textbf{V-Measure} \cite{rosenberg2007v} (standing for validity) is the harmonic mean between homogeneity and completeness:
    
    $$ v = \frac{(1+\beta) h c}{\beta h + c} $$
    In practice, $\beta=1$ is used to weight $h$ and $c$ equally. 
    
    \item \textbf{Adjusted Rand Index} \cite{hubert1985comparing} is a version of the Rand index, a measure of cluster similarity, which is adjusted for chance. 
    
    $$ ARI = \frac{RI - \mathbb{E} \left[ RI \right] }{\max RI - \mathbb{E} \left[ RI \right] } $$
    where the Rand index, $RI$, is 
    
    $$ RI = \frac{TP + TN}{{n \choose 2}} $$
    and $\mathbb{E} \left[ RI \right]$ is expected value of random clusterings. 
\end{itemize}

%% file: appendixH.tex
\section{UMAP Weights}
\label{appendix:umap}

UMAP \cite{mcinnes2018umap} attempts to estimate the density by comparing the distance to the k-nearest neighbors as follows: 

$$\rho_i = min\{d(x_i, x_{i_j})|1\le j \le k, d(x_i, x_{i_j}) > 0\}$$

$$\sum_{j=1}^k\exp(\frac{-max(0, d(x_i, x_{i_j})-\rho_i)}{\sigma_i}) = \log_2(k)$$

Here, $d(x_i, x_{i_j})$ is the distance between $x_i$ and $x_{i_j}$. $\rho_i$ is the minimum distance to $x_i$'s closest neighbor. $\sigma_i$, which smooths and normalizes the distances to the nearest neighbors, is calculated for each data point. Next, UMAP calculates the following weights between data points:

$$w((x_i, x_j)) = \exp(\frac{-max(0, d(x_i, x_{i_j})-\rho_i)}{\sigma_i})$$

We use $1 - w((x_i, x_j))$ for agglomerative clustering.

%% file: appendixI.tex
\section{Reproducibility}
\label{appendix:reproducibility}

The SBERT model we used, along with the size of the $Q$ and $K$ layers use a dimension of size 384. Our total model has 34,839,937 parameters, of which 1,479,937 do not belong to SBERT. Input uses the `ldc\_scope' part of the ACE event mention. Our model takes roughly 2 hours to train on one V100 GPU, including the early stopping calculations which are done with the model set to `evaluation' mode. We used batch size 10, which is the most that would fit in memory. For learning rates, we considered the suggestions in \cite{devlin2018bert}.